\documentclass{article}

\PassOptionsToPackage{numbers, compress}{natbib}

\usepackage[preprint]{neurips_2024}
\usepackage{graphicx}    
\usepackage{amsmath}     
\usepackage{algorithm}   
\usepackage{algpseudocode}
\usepackage{subfigure}   
\usepackage{booktabs}    
\usepackage{xcolor}      
\usepackage{amsfonts}    
\usepackage{nicefrac}    
\usepackage{hyperref}    
\usepackage{url}         
\usepackage{wrapfig}     
\usepackage{multirow}    
\usepackage{newunicodechar} 
\usepackage[capitalize,noabbrev]{cleveref} 
\usepackage{enumitem}
\usepackage{multicol}
\usepackage{adjustbox}
 \usepackage{mathrsfs}

\usepackage[utf8]{inputenc} 
\usepackage[T1]{fontenc}    
\usepackage{hyperref}       
\usepackage{url}            
\usepackage{booktabs}       
\usepackage{amsfonts}       
\usepackage{nicefrac}       
\usepackage{microtype}      
\usepackage{xcolor}         

\title{Biomedical SAM-2: \\Segment Anything in Biomedical Images and Videos}

\author{
  $\text{\textbf{Zhiling Yan}}^{\textbf{1}\thanks{Equal Contributions}}$~~~~~~~~~~$\text{\textbf{Weixiang Sun}}^{\textbf{2}*}$~~~~~~~~~~$\text{\textbf{Rong Zhou}}^{\textbf{1}*}$~~~~~~~~~~$\text{\textbf{Zhengqing Yuan}}^{\textbf{3}*}$\\
  $\text{\textbf{Kai Zhang}}^{\textbf{1}}$~~~~~~~~~~$\text{\textbf{Yiwei Li}}^{\textbf{4}}$~~~~~~~~~~$\text{\textbf{Sekeun Kim}}^{\textbf{5}}$~~~~~~~~~~$\text{\textbf{Tianming Liu}}^{\textbf{4}}$~~~~~~~~~~$\text{\textbf{Quanzheng Li}}^{\textbf{5}}$\\
  $\text{\textbf{Xiang Li}}^{\textbf{5}}$~~~~~~~~~~$\text{\textbf{Lifang He}}^{\textbf{1}}$~~~~~~~~~~$\text{\textbf{Lichao Sun}}^{\textbf{1}}\thanks{Corresponding Author: lis221@lehigh.edu}$\\
  ${}^{\textbf{1}}$Lehigh University, 
  ${}^{\textbf{2}}$Northeastern University, 
  ${}^{\textbf{3}}$University of Notre Dame, \\
  ${}^{\textbf{4}}$University of Georgia, 
  ${}^{\textbf{5}}$Massachusetts General Hospital, Harvard Medical School\\
}

\begin{document}

\maketitle

\begin{abstract}

Medical image segmentation and video object segmentation are essential for diagnosing and analyzing diseases by identifying and measuring biological structures. Recent advances in natural domain have been driven by foundation models like the Segment Anything Model 2 (SAM-2). To explore the performance of SAM-2 in biomedical applications, we designed three evaluation pipelines for single-frame 2D image segmentation, multi-frame 3D image segmentation and multi-frame video segmentation with varied prompt designs, revealing SAM-2's limitations in medical contexts. Consequently, we developed BioSAM-2, an enhanced foundation model optimized for biomedical data based on SAM-2. Our experiments show that BioSAM-2 not only surpasses the performance of existing state-of-the-art foundation models but also matches or even exceeds specialist models, demonstrating its efficacy and potential in the medical domain. 

\begin{figure}[H]
    \centering
    \includegraphics[width=.9\linewidth]{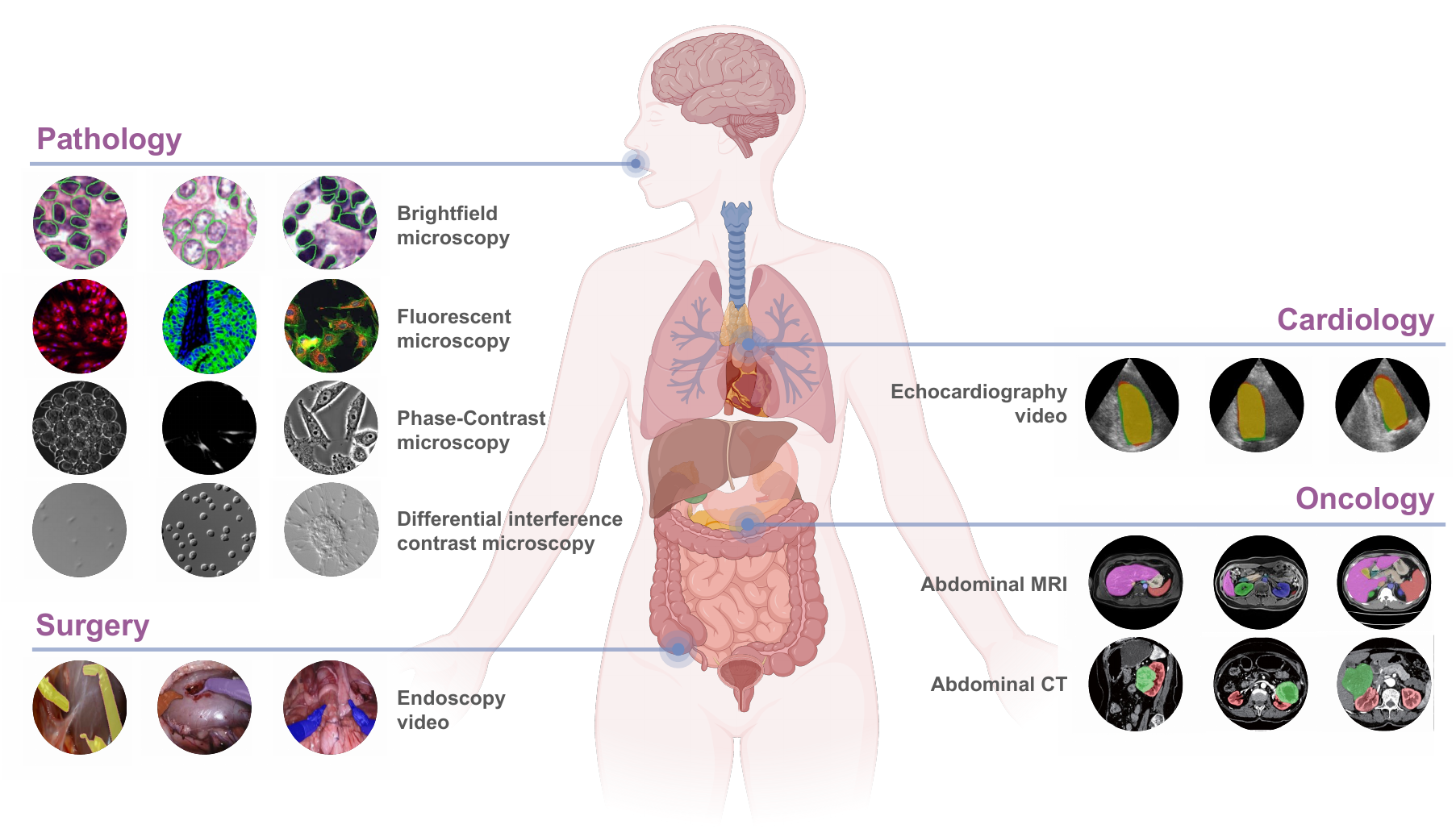}
    \caption{Overview of biomedical tasks in BioSAM-2.}
    \label{fig:overall}
\end{figure}
\end{abstract}

\section{Introduction}

Medical image segmentation is crucial for identifying biological structures and measuring their morphology, aiding in the diagnosis and analysis of various diseases \cite{norouzi2014medical, du2020medical}. 
Despite numerous advancements in medical imaging technologies, segmentation remains a formidable challenge due to the complexity of medical images and the extensive manual effort required for accurate annotation \cite{mazurowski2023segment, ramesh2021review}. 
Traditional methods often necessitate detailed manual annotation, which is not only time-consuming but also prone to human error \cite{yin2018medical}. 

Recently, the emergence of segmentation foundation models, such as the Segment Anything Model (SAM) \cite{kirillov2023segment} has driven significant advancements in the field of natural image segmentation.
SAM demonstrates impressive zero-shot segmentation performance with prompt inputs, showing remarkable versatility and setting a new standard across various segmentation tasks. To further adapt SAM's capabilities to the medical field, numerous works have been proposed~\cite{ma2024segment, chen2023masammodalityagnosticsamadaptation, zhang2023customizedsegmentmodelmedical, dai2024samaugpointpromptaugmentation}, with MedSAM~\cite{ma2024segment} being a representative one. MedSAM modifies the SAM architecture by incorporating domain-specific knowledge to address the unique challenges of medical images, such as varying contrast, noise levels, and the presence of artifacts~\cite{9103969}. This approach has demonstrated significant improvements in segmentation performance on medical images, leveraging SAM's foundation while tailoring it for medical applications. 

Recognizing the need to extend these capabilities to more complex scenarios, SAM-2 \cite{ravi2024sam2} was developed to expand SAM's functionality to include video inputs. This extension enables SAM-2 to process temporal sequences of images, making it suitable for tasks that require understanding spatial continuity over multiple frames. By handling both spatial and temporal dimensions, SAM-2 has demonstrated impressive zero-shot performance on various tasks involving natural image and video segmentation.

However, SAM-2's potential on medical segmentation tasks has not yet been fully explored. We conducted a comprehensive evaluation to investigate its capability. Specifically, we assessed the performance of four SAM-2 variants (Hiera-T, Hiera-S, Hiera-B+, and Hiera-L) across 8 medical modalities and 22 objects of interest. We designed three evaluation pipelines for single-frame 2D image segmentation, multi-frame 3D image segmentation and multi-frame video segmentation, respectively, incorporating diverse prompt designs. To further evaluate its performance, we compared it against various baseline models, including CNN-based, Transformer-based, and SSM-based models, using a range of metrics. Our findings indicate that SAM-2 cannot be directly utilized for medical image or video segmentation. The main reasons are the significant domain gap between natural and medical data and its inability to associate segmentation regions with meaningful semantic classes. In other words, SAM-2 cannot perform semantic segmentation in the medical domain, which limits its application in computer-aided diagnosis.

Building on these observations, we further introduce BioSAM-2, a refined foundation model that significantly enhances the segmentation performance of SAM-2 on biomedical images and videos. BioSAM-2 incorporates a memory mechanism and a stream processing architecture, which is the sam as SAM-2. This model can handle multi-frame segmentation tasks by retaining information from past predictions, allowing it to make accurate predictions on slices without explicit prompts. Experimental results demonstrate that BioSAM-2 consistently outperforms the state-of-the-art (SOTA) segmentation foundation model~\cite{kirillov2023segment, ravi2024sam2}, while on par with, or even surpassing the performance of specialist models~\cite{isensee2021nnu, myronenko20193d, hatamizadeh2022unetr, hatamizadeh2022swinunetrswintransformers} trained on medical data from the same modality. These findings underscore the potential of BioSAM-2 as a new paradigm for versatile medical image and video segmentation.


\begin{figure}[h]
    \centering
\includegraphics[width=1\linewidth]{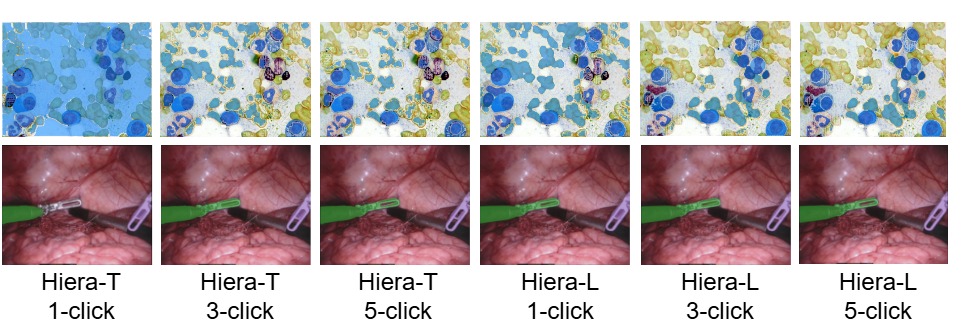}
    \caption{Image segmentation results of tiny SAM-2 and large SAM-2 based on different segmentation prompts.
}
    \label{fig:image_visualization}
\end{figure}

Our contributions can be summarized as follows:
\begin{itemize}[noitemsep,topsep=0pt]
    \item We have developed three evaluation pipelines tailored for single-frame biomedical images, multi-frame biomedical images and multi-frame biomedical videos within 8 medical modalities and 22 objects of interest. These pipelines comprehensively assess the performance of SAM-2 in biomedical applications.

    \item To enhance the adaptability of SAM-2 in the biomedical domain, we introduce BioSAM-2, an optimized foundational model achieved by fine-tuning the original SAM-2. This refinement significantly boosts the segmentation performance of SAM-2. Remarkably, without any prompts, our automated segmentation consistently outperforms competitive SOTA foundation methods by a large margin.

    \item The results of experiments show that BioSAM-2 matches or even exceeds the performance of specialized models trained on medical data of the same modality. These findings highlight the potential of BioSAM-2 as a novel paradigm for versatile medical applications.
    
\end{itemize}

By integrating BioSAM-2 with medical image segmentation tasks, we anticipate significant improvements in segmentation accuracy and annotation efficiency, ultimately contributing to better clinical outcomes and facilitating the adoption of AI in medical imaging. This research aims to push the boundaries of current medical image segmentation techniques and explore the full potential of advanced AI models like BioSAM-2 in handling the complexities of medical imaging data.


\section{Related Work}
\textbf{Medical Image Segmentation.}~~CNN-based and Transformer-based models have significantly advanced medical image segmentation. U-Net~\cite{ronneberger2015u}, a notable CNN-based approach, features a symmetrical encoder-decoder architecture with skip connections to preserve details. Enhancements~\cite{myronenko20193d} like the self-configuring nnU-Net~\cite{isensee2021nnu} have demonstrated robust performance across various medical segmentation challenges. In Transformer-based models, TransUnet~\cite{chen2021transunet} integrates the Vision Transformer (ViT)\cite{dosovitskiy2020image} for feature extraction and pairs it with a CNN for decoding, effectively processing global information. UNETR~\cite{hatamizadeh2022unetr}, and Swin-UNet~\cite{cao2022swin} combine Transformer architectures with U-Net to enhance 3D imaging analysis, exploring Swin Vision Transformer blocks~\cite{liu2021swin}.SSM-based models like U-Mamba~\cite{ma2024u} have also been introduced for efficient long-sequence data analysis in medical imaging. Recently, SAM~\cite{kirillov2023segment}, a vision foundation model pre-trained on 1 billion masks, demonstrated impressive zero-shot capability across various segmentation tasks. Inspired by SAM's performance in natural images, adaptations quickly emerged for medical segmentation~\cite{zhang2023customized, biswas2023polyp, wu2023self, dai2023samaug}. MedSAM~\cite{ma2024segment} fine-tuned SAM on over 200,000 masks across 11 modalities, while SAM-Med2D~\cite{cheng2023sam} used comprehensive prompts for 2D medical images. SAMed~\cite{zhang2023customized} and MA-SAM~\cite{chen2023ma} employed PETL~\cite{hu2021lora} for fine-tuning, outperforming several existing medical segmentation methods.

\textbf{Medical Video Object Segmentation.}~~A large number of semantic segmentation models rely on single images to identify objects in a scene. This can lead to spatially and temporally inconsistent predictions especially in multi-frame videos that require temporal context. To address this, Space Time Memory Networks (STM)~\cite{oh2019videoobjectsegmentationusing} and its variants~\cite{cheng2021rethinkingspacetimenetworksimproved, 9879303, liang2020videoobjectsegmentationadaptive} use a memory network to extract vital information from a time-based buffer composed of all previous video sequences. Building on this methodology, DPSTT~\cite{10.1007/978-3-031-16440-8_38} integrates a memory bank with decoupled transformers to track temporal lesion movement in medical ultrasound videos. However, DPSTT requires substantial data augmentation to avoid overfitting and suffers from low processing speed. Subsequently, FLANet~\cite{lin2023shiftingattentionbreastlesion} introduced a frequency and location feature aggregation network, involving a large amount of memory occupancy. 
Optical flow methods for surgical videos~\cite{gonzález2020isinetinstancebasedapproachsurgical, zhao2020learningmotionflowssemisupervised} are limited to using features between pairs of images and cannot leverage extended temporal context. Other methods employ a combination of 2D encoders and 3D convolutional layers in the temporal decoder~\cite{10.1007/978-3-030-59725-2_29} and Convolutional Long short-term Memory cells~\cite{wang2020noisylstmimprovingtemporalawareness}. Alternative approaches enforce temporal consistency through a loss function during training~\cite{liu2020efficientsemanticvideosegmentation} or use architectures that combine high and low frame rate model branches to integrate temporal context from different parts of the video~\cite{jain2019accelcorrectivefusionnetwork}. 

The recently introduced SAM-2~\cite{ravi2024sam2} extended the backbone of SAM to 3D, enhancing its capability to ``segment anything in videos''. Specifically, SAM-2 is equipped with a memory that stores information about objects and previous interactions, allowing it to generate and correct masklet predictions throughout the video.

\section{Method}

\subsection{Preliminary Study of SAM-2}
Segment Anything Model 2 (SAM-2) is a unified transformer-based model for both image and video segmentation. 
For each video frame, the segmentation prediction leverages the current prompt and previously observed memories. Videos are processed sequentially, with each frame handled individually by \textbf{image encoder}, while \textbf{memory attention} conditions current frame features on past frames and predictions. The \textbf{mask decoder}, which can optionally take input prompts, predicts the segmentation mask for that frame. Finally, a \textbf{memory encoder} transforms the predictions and image embeddings into a form usable for future frames, ensuring temporal consistency among multiple frames.

The vision transformer in the image encoder is pretrained using the hierarchical masked autoencoder model Hiera~\cite{ryali2023hierahierarchicalvisiontransformer}, enabling multiscale feature decoding. Memory attention conditions the current frame features on past frames' features and predictions. Multiple transformer blocks are stacked, with the first block taking the current frame's image encoding as input. Each block performs self-attention, followed by cross-attention to memories of frames and object pointers stored in a memory bank. In SAM-2, prompts are encoded with positional encoding and two learnable tokens specifying foreground and background. The mask decoder comprises bi-directional transformer blocks that update prompt and frame embeddings. The model predicts multiple masks per frame, and in cases of ambiguity without subsequent prompt clarification, it propagates only the mask with the highest predicted IoU. Additionally, an auxiliary prediction head determines the presence of the target object in the current frame. Finally, the memory encoder consolidates this process by downsampling the output mask using a convolutional module, which is then element-wise summed with the unconditioned frame embedding from the image encoder. This integrated data is retained in a memory bank, preserving essential information about past predictions for the target object throughout the video sequence.

\begin{figure}[h]
    \centering
\includegraphics[width=.9\linewidth]{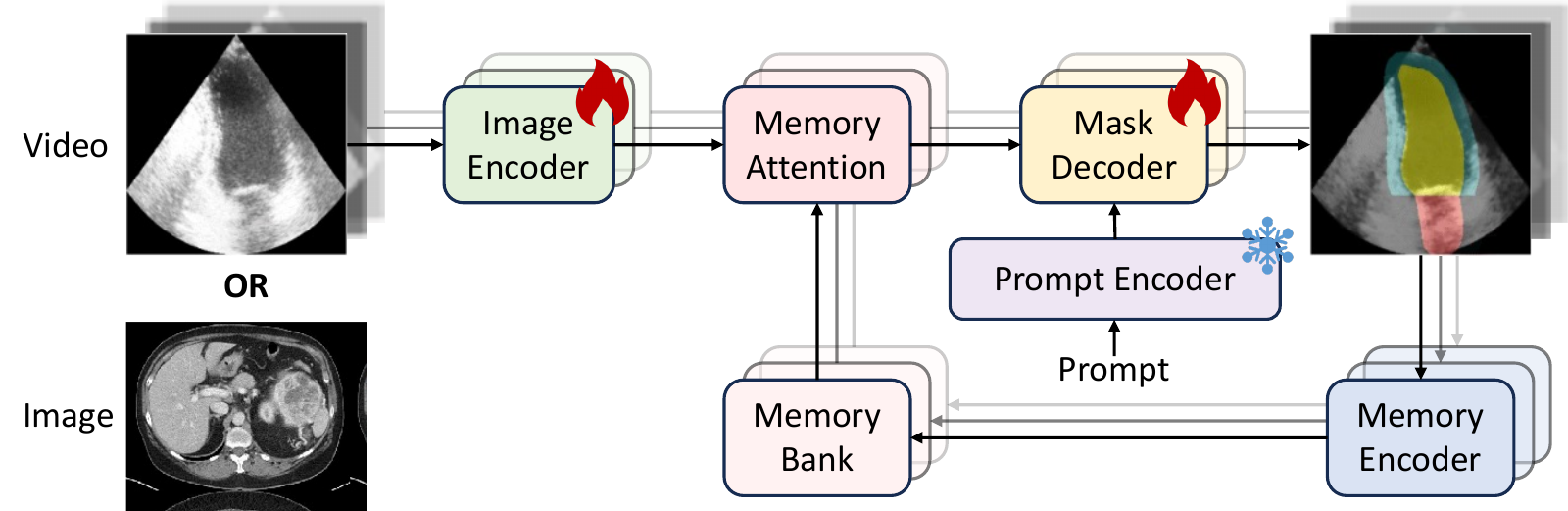}
    \caption{The workflow of proposed BioSAM-2. We adapt SAM-2 for medical image and video segmentation by freezing the prompt encoder and only finetuning the image encoder and mask decoder.}
    \label{fig:architecture}
\end{figure}

\subsection{Medical Applications of SAM-2}

Building on the impressive zero-shot capabilities demonstrated by SAM-2 in natural images and videos, we explore its performance in medical applications. Specifically, we designed three pipelines, single-frame image segmentation, multi-frame iamge segmentation and multi-frame video segmentation to assess SAM-2's ability to handle zero-shot segmentation tasks within these medical contexts.

\textbf{Single-frame 2D Image Segmentation.} 
We design single-frame image segmentation in a non-iterative manner in which all prompts are determined without feedback from any prior predictions. 

Firstly, we automate the generation of point prompts through a specific strategy. One point is added as the prompt by randomly selecting it from the initial mask. Given a set of candidate points $P$ derived from the initial mask, the selected point $p$ is: 
\begin{equation}
p = random(P)
\end{equation}

As highlighted in the SAM-2 documentation~\cite{ravi2024sam2}, employing a single point prompt can lead to segmentation ambiguity, as the model may associate the prompt with multiple valid masks without being able to discern the intended one. Although SAM-2 incorporates an ambiguity-resolving module that generates multiple masks and ranks them based on confidence scores, the use of multiple point prompts significantly mitigates this issue. Thus, we also assessed the performance of SAM-2 with additional randomly selected points from the initial mask candidates.

This sampling-from-mask method treats the initial segmentation mask as a reliable outcome and seeks to enhance segmentation accuracy by exploiting prompt selection invariance and incorporating additional point prompts.

\textbf{Multi-frame 3D Image Segmentation.}
For multi-frame 3D image segmentation, we initially set the bounding box prompts on the middle slice of the image volume, following~\cite{ma2024segmentmedicalimagesvideos, dong2024segmentmodel2application}. The middle slice is chosen as the starting point in 3D image segmentation due to its tendency to contain the largest object and the greatest number of semantic classes among all slices in the conventional axial view. This strategic choice enhances the accuracy and reliability of subsequent segmentation. Once the middle frame is annotated, SAM-2 facilitates the propagation of annotations to the surrounding slices that do not have prompts. This is achieved through the video segmentation feature integrated into SAM-2. The process involves first propagating the annotations backward from the middle slice to the first slice, followed by a forward propagation from the middle slice to the last slice in the volume. This bidirectional propagation ensures consistent segmentation across the entire image volume.

\textbf{Multi-frame Video Segmentation.}
In evaluating video segmentation, we employ a hybrid mode of offline and online evaluation. Specifically, we select the first n frames as interacted frames, on which click cues are added. Multiple click cues enable the model to more accurately determine the boundaries of objects, achieving higher segmentation accuracy. Additionally, multiple interacted frames effectively resolve the issue identified in SAM-2 where the model fails to track objects appearing in unmarked frames, making it more reasonable and effective for video scenarios. Overall, our evaluation traverses the video only once, subsequently yielding the final segmentation results.

To summarize, when directly employing SAM-2 for medical image and video segmentation, the generated mask can be ambiguous and necessitates multiple prompts or iterations for prediction and correction. Despite its promising potential, SAM-2 has encountered challenges in delivering satisfactory segmentation results across various medical image/video segmentation tasks. Additionally, since the video training data for SAM-2 predominantly consists of high-resolution footage, it may become entirely ineffective when dealing with low-resolution medical videos. In light of these limitations, the objective of this study is to develop a robust segmentation foundation model capable of effectively addressing a wide range of biomedical segmentation targets.

\subsection{BioSAM-2: Dedicated biomedical segmentation foundation model}

To tailor SAM-2 for medical video segmentation, selecting the appropriate network components for fine-tuning is crucial. The architecture of SAM-2 comprises several key elements: image encoder, prompt encoder, memory attention, mask decoder, and memory encoder. It is possible to fine-tune any combination of these components. For our adaptation, the prompt encoder, which processes the information from the given prompts, is retained from the pre-trained model and is therefore frozen to maintain its original functionality. Conversely, the image encoder and mask decoder are fine-tuned to enhance SAM-2's suitability for the medical imaging domain. This adaptation strategy is depicted in Figure~\ref{fig:architecture}.

For the image encoder, we opted for a tiny-sized configuration to strike a balance between computational cost and performance efficacy. We initiated training for SAM-2 from the official checkpoint, conducting separate sessions for image and video data. During training, we employed the AdamW optimizer~\cite{loshchilov2017decoupled} and implemented layer decay strategies~\cite{clark2020electra} on the image encoder to enhance its learning efficiency. In terms of the mask decoder, its configuration is simplified to generate a single mask per frame, given that the provided prompt distinctly identifies the expected segmentation target.

For loss design, we supervise the model’s predictions using a combination of dice loss and binary cross-entropy (BCE) loss for the mask prediction. Specifically, $p_i$ and $g_i$ are the predicted and ground truth pixel values, respectively, and $N$ is the total number of pixels. The dice loss is defined as:
\begin{equation}
\mathcal{L}_{\text{dice}} = 1 - \frac{2 \times \sum_{i=1}^N p_i g_i}{\sum_{i=1}^N p_i + \sum_{i=1}^N g_i}
\end{equation}
and BCE loss is:
\begin{equation}
\mathcal{L}_{\text{BCE}} = -\frac{1}{N} \sum_{i=1}^N \left[ g_i \log(p_i) + (1 - g_i) \log(1 - p_i) \right]
\end{equation}

During training, the model parameters are optimized using a combination of these losses:
\begin{equation}
\mathcal{L} = \alpha\mathcal{L}_{\text{dice}} +  \beta\mathcal{L}_{\text{BCE}}
\end{equation}
If the ground-truth does not contain a mask for a frame, we do not supervise any of the mask outputs, but always supervise the occlusion prediction head that predicts whether there should exist a mask in the frame. 

\section{Experiments}

\subsection{Biomedical Image Segmentation}

\subsubsection{Datasets}
To assess the performance and scalability of BioSAM-2, we utilize four medical image datasets, including Abdomen CT dataset~\cite{ma2023unleashing}, Abdomen MR dataset~\cite{ji2022amos}, Endoscopy dataset~\cite{allan20192017} and Microscopy dataset~\cite{ma2023multi}.

\textbf{Abdomen 3D images.}~~For evaluating BioSAM-2, two datasets are implemented. The first is Abdomen CT~\cite{ma2023unleashing} from the MICCAI 2022 FLARE challenge, which includes segmentation of 13 abdominal organs from 50 CT scans in both the training and testing sets. The organs include the liver, spleen, pancreas, kidneys, stomach, gallbladder, esophagus, aorta, inferior vena cava, adrenal glands, and duodenum. The second dataset, Abdomen MR~\cite{ji2022amos} from the MICCAI 2022 AMOS Challenge, involves the same 13 organs with 60 MRI scans for training and 50 for testing.

\textbf{Endoscopy Images.}~~From the MICCAI 2017 EndoVis Challenge~\cite{allan20192017}, this dataset focuses on instrument segmentation within endoscopy images, featuring seven distinct instruments, including the large needle driver, prograsp forceps, monopolar curved scissors, cadiere forceps, bipolar forceps, vessel sealer, and a drop-in ultrasound probe. The dataset is split into 1800 training frames and 1200 testing frames.

\textbf{Microscopy images.}~~This dataset, from the NeurIPS 2022 Cell Segmentation Challenge~\cite{ma2023multi}, is used for cell segmentation in microscopy images, consisting of 1000 training images and 101 testing images. Following U-Mamba~\cite{ma2024u}, we address this as a semantic segmentation task, focusing on cell boundaries and interiors rather than instance segmentation.

\subsubsection{Experimental Setup}
The setting of our experiments is the same as that in the corresponding official repository for each method to ensure a fair comparison. 
We adopt an unweighted combination of Dice loss and cross-entropy loss for all datasets and utilize the AdamW optimizer with an initial learning rate of 1e-4. The training duration for each dataset is set to 200 epochs, with batch size as 8. For the evaluation of SAM-based methods, we follow the implementation details of their official technique reports. We introduce different prompt designs and feed them into the model. All prompts are randomly chosen from the corresponding mask of the image. 

In our evaluation of BioSAM-2, we compare against two prominent CNN-based segmentation methods: nnUNet~\cite{isensee2021nnu} and SegResNet~\cite{myronenko20193d}. Additionally, we include a comparison with UNETR~\cite{hatamizadeh2022unetr} and SwinUNETR~\cite{hatamizadeh2022swinunetrswintransformers}, two Transformer-based networks that have gained popularity in medical image segmentation tasks. U-Mamba~\cite{ma2024u}, a recent method based on the Mamba model, is also included in our comparison to provide a comprehensive overview of its performance. For each model, we implement their recommended optimizers to ensure consistency in training conditions. To maintain fairness across all comparisons, we finetune all these models on each dataset and apply the default image preprocessing in nnUNet~\cite{isensee2021nnu}. We also assess the performance of SAM and SAM-2 by directly allowing them to infer the corresponding mask for the image. To ensure a thorough evaluation, we test two sizes of SAM-2 using three kinds of prompt.

For evaluation metrics, we employ the Dice Similarity Coefficient (DSC) and Normalized Surface Distance (NSD) to assess performance in abdominal multi-organ segmentation for CT and MR scans, as well as instrument segmentation in endoscopy images. For the cell segmentation task, we utilize the F1 score and NSD to evaluate method performance.

\begin{table}[h]
\small
\caption{Results summary of 2D image segmentation tasks: instruments segmentation in endoscopy images, and cell segmentation in microscopy images.}\label{tab:results-2d}
\centering
\begin{adjustbox}{width=0.99\textwidth}
\begin{tabular}{lll|cc}
\toprule
\multirow{2}{*}{Methods}  & \multicolumn{2}{c|}{Instruments in Endoscopy}           & \multicolumn{2}{c}{Cells in Microscopy}           \\ \cline{2-5}                        & \multicolumn{1}{c}{DSC}    & \multicolumn{1}{c|}{NSD}   & \multicolumn{1}{c}{F1}    & \multicolumn{1}{c}{NSD}                        \\ \hline
nnU-Net                        & 0.6264\tiny{$\pm$0.3024}          & 0.6412\tiny{$\pm$0.3074}          & 0.5383\tiny{$\pm$0.2657}      &    \textbf{0.8332\tiny{$\pm$0.1611}}\\
SegResNet                       & 0.5820\tiny{$\pm$0.3268}          & 0.5968\tiny{$\pm$0.3303}          & 0.5411\tiny{$\pm$0.2633}     & 
0.7944\tiny{$\pm$0.2356}     \\
UNETR                             & 0.5017\tiny{$\pm$0.3201}          & 0.5168\tiny{$\pm$0.3235}          & 0.4357\tiny{$\pm$0.2572}    & 0.8201\tiny{$\pm$0.2263}      \\
SwinUNETR                      & 0.5811\tiny{$\pm$0.3176}          & 0.5973\tiny{$\pm$0.3209}          & 0.3880\tiny{$\pm$0.2664}     &  0.7981\tiny{$\pm$0.1920}  \\
U-Mamba                      & \textbf{0.6303\tiny{$\pm$0.3067}}          & \textbf{0.6451\tiny{$\pm$0.3104}}         & 0.5607\tiny{$\pm$0.2784}     & 0.8288\tiny{$\pm$0.1706}   \\ \hline
SAM  & 0.1583\tiny{$\pm$0.3300}          & 0.1600\tiny{$\pm$0.3313}          & 0.3249\tiny{$\pm$0.2285} & 
0.3696\tiny{$\pm$0.2544} \\
SAM-2 (Hiera-T, 1-click)  &
0.4115\tiny{$\pm$0.4092} &
0.4227\tiny{$\pm$0.4189} & 0.0654\tiny{$\pm$0.1220} &
0.0720\tiny{$\pm$0.1372}\\
SAM-2 (Hiera-T, 3-click)  &
0.5215\tiny{$\pm$0.3802} &
0.5349\tiny{$\pm$0.3864} & 0.3436\tiny{$\pm$0.2400} & 0.3911\tiny{$\pm$0.2640} \\
SAM-2 (Hiera-T, 5-click)  &
0.5382\tiny{$\pm$0.3568} &
0.5520\tiny{$\pm$0.3616} & 0.3566\tiny{$\pm$0.2496} & 0.4070\tiny{$\pm$0.2761} \\
SAM-2 (Hiera-L, 1-click) & 0.4416\tiny{$\pm$0.4217} & 0.4523\tiny{$\pm$0.4312} & 0.0799\tiny{$\pm$0.1540} & 0.0877\tiny{$\pm$0.1722} \\
SAM-2 (Hiera-L, 3-click) & 0.5354\tiny{$\pm$0.3750} & 0.5497\tiny{$\pm$0.3813} & 0.3217\tiny{$\pm$0.2481} & 0.3720\tiny{$\pm$0.2750} \\
SAM-2 (Hiera-L, 5-click) & 0.5479\tiny{$\pm$0.3629} & 0.5623\tiny{$\pm$0.3681} & 0.3352\tiny{$\pm$0.2598} & 0.3876\tiny{$\pm$0.2849} \\
BioSAM-2 (Hiera-T) & 0.6251\tiny{$\pm$0.2897} & 0.6427\tiny{$\pm$0.3095} & \textbf{0.5792\tiny{$\pm$0.2666}} & 0.7436\tiny{$\pm$0.2104}\\
\hline
\end{tabular}
\end{adjustbox}
\end{table}

\subsubsection{Results}

As illustrated in Table~\ref{tab:results-2d} and Table~\ref{tab:results-3d}, we conducted a thorough analysis of various SAM-2 variants. The results indicate a clear performance improvement with an increased number of clicks, particularly noticeable in microscopy datasets where the F1 metric improved from 0.0654 to a maximum of 0.3566. Similarly, significant progress was observed in other datasets. For instance, the DSC score of endoscopy dataset rises from a minimum of 0.4115 to 0.5382, and the NSD score increases from 0.4227 to 0.5520. Furthermore, we evaluated SAM-2's performance across two different model sizes. The larger model size outperformed the tiny variant in terms of majority DSC and NSD scores, given the same prompt. This suggests that a larger SAM-2 model possesses superior segmentation capabilities.

When comparing the results of zero-shot of SAM-2 to other fine-tuned models specifically designed for medical image segmentation (e.g. nnU-Net), it is evident that the zero-shot performance is inferior. This discrepancy highlights that despite SAM2's robust transfer learning capabilities, there remains significant room for improvement in the medical imaging domain. This underscores the necessity of fine-tuning of SAM-2 to achieve optimal performance. In addition, it is observed that when SAM-2 employs proper prompts, SAM-2's zero-shot results outperform those of SAM, even though SAM utilizes its largest version, SAM\_h. It reinforces the advantage of SAM-2's advanced design and adaptability, compared with SAM.

Table~\ref{tab:results-2d} and Table~\ref{tab:results-3d} also show the performance of our proposed method BioSAM-2. Comparative analysis between BioSAM-2 and SAM-2 reveals substantial improvements with BioSAM-2, achieving enhancements ranging from a minimum of 0.0772 to a maximum of 0.5138. This shows BioSAM-2's superior performance on biomedical imaging segmentation and demonstrates SAM-2's significant potential in this domain. SAM-2, being a universal model, requires adaptation, particularly in the medical field, due to its general-purpose design rather than being specialized. Its limited knowledge base on medical datasets and the constrained number of output masks are key factors. While SAM-2 can effectively segment image-level instances, it struggles with accurately segmenting class-level instances. For example, SAM-2 can easily delineate the boundaries of two cells but cannot determine if they belong to the same class. These limitations impact SAM-2's performance, especially in multi-class medical segmentation datasets. BioSAM-2 thus plays a critical role in bridging these gaps, enhancing the model's capability to handle the intricacies of medical image segmentation tasks.

Finally, according to Table~\ref{tab:results-2d}, BioSAM-2 achieved a DSC score of 0.6251 and an NSD score of 0.6427 on the endoscopy dataset. On the microscopy dataset, it obtained an F1 score of 0.5792 and an NSD score of 0.7436. These results surpass most competing methods and are comparable to state-of-the-art models. This performance demonstrates BioSAM-2's exceptional capability in medical image segmentation, underscoring its potential to deliver high-quality results.


\begin{table}[htb]
\small
\caption{Results summary of 3D organ segmentation on abdomen CT and MR datasets.}\label{tab:results-3d}
\centering
\begin{adjustbox}{width=0.99\textwidth}
\begin{tabular}{lcc|cc}
\toprule
\multirow{2}{*}{Methods} & \multicolumn{2}{c|}{Organs in Abdomen CT}                 & \multicolumn{2}{c}{Organs in Abdomen MR}                 \\ \cline{2-5} 
                         & DSC                    & NSD                    & DSC                    & NSD                    \\ \hline
nnU-Net                  & 0.8615\tiny{$\pm$0.0790}          & 0.8972\tiny{$\pm$0.0824}          & 0.8309\tiny{$\pm$0.0769}          & 0.8996\tiny{$\pm$0.0729}          \\
SegResNet                & 0.7927\tiny{$\pm$0.1162}         & 0.8257\tiny{$\pm$0.1194}          & 0.8146\tiny{$\pm$0.0959}          & 0.8841\tiny{$\pm$0.0917}          \\
UNETR                    & 0.6824\tiny{$\pm$0.1506}          & 0.7004\tiny{$\pm$0.1577}          & 0.6867\tiny{$\pm$0.1488}          & 0.7440\tiny{$\pm$0.1627}          \\
SwinUNETR                & 0.7594\tiny{$\pm$0.1095} & 0.7663\tiny{$\pm$0.1190} & 0.7565\tiny{$\pm$0.1394}          & 0.8218\tiny{$\pm$0.1409}  \\

U-Mamba             & \textbf{0.8638\tiny{$\pm$0.0908}} & \textbf{0.8980\tiny{$\pm$0.0921}} & \textbf{0.8501\tiny{$\pm$0.0732}} & \textbf{0.9171\tiny{$\pm$0.0689}} \\ \hline
SAM & 0.2957\tiny{$\pm$0.2761} & 0.2967\tiny{$\pm$0.3018} & 0.3710\tiny{$\pm$0.4321} & 0.3712\tiny{$\pm$0.4321} \\

SAM-2 (Hiera-T) & 0.4431\tiny{$\pm$0.2690} & 0.3875\tiny{$\pm$0.2544}  & 0.5534\tiny{$\pm$0.2609}  & 0.5572\tiny{$\pm$0.2683}     \\
SAM-2 (Hiera-L) & 0.4744\tiny{$\pm$0.2472} & 0.4208\tiny{$\pm$0.2306}  & 0.5492\tiny{$\pm$0.2315} & 0.5526\tiny{$\pm$0.2341}     \\
BioSAM-2 (Hiera-T) & 0.7632\tiny{$\pm$0.0836} & 0.8568\tiny{$\pm$0.0732} & 0.7439\tiny{$\pm$0.0629} & 0.8317\tiny{$\pm$0.0746} \\
\hline
\end{tabular}
\end{adjustbox}
\end{table}

\subsection{Biomedical Video Segmentation}
\subsubsection{Datasets}
To validate the performance of SAM-2 on biomedical video, we selected two datasets from medical scenarios, which include EndoVis 2018~\cite{allan20202018} and EchoNet-Dynamic~\cite{ouyang2020video}.

\textbf{EndoVis 2018.}~~From Robotic Scene Segmentation Challenge~\cite{allan20202018}. This dataset consists of video data from 16 robotic nephrectomy procedures performed using da Vinci Xi systems in porcine labs, aimed at supporting machine learning research in surgical robotics. Originally recorded at a high frequency of 60 Hz, the data has been subsampled to 2 Hz to manage labeling costs, resulting in 149 frames per procedure after removing sequences with minimal motion. Each frame, presented in a resolution of 1280$\times$1024, includes images from both left and right eye cameras along with stereo camera calibration parameters.

\textbf{EchoNet-Dynamic.}~~The EchoNet-Dynamic dataset~\cite{Ouyang2020} featuring 10,030 labeled echocardiogram videos was collected from routine clinical care at Stanford University Hospital between 2016 and 2018. This extensive dataset provides a unique resource for studying cardiac motion and chamber sizes, which is crucial for diagnosing a range of cardiovascular diseases. Each video in the dataset captures the heart's dynamics from the apical-4-chamber view, meticulously cropped and masked to eliminate any extraneous text and external information, ensuring a focus solely on the cardiac imaging area. The videos have been uniformly resized to 112$\times$112 pixels using cubic interpolation to standardize the dataset.

\begin{table}[t]
    \centering
    \small
    \caption{Different-sized models' $\mathcal{J\&F}$ scores ($\%$) under zero-shot conditions.}
    \begin{adjustbox}{width=0.99\textwidth}
    \begin{tabular}{lcccc|cccc}
    \hline
        \multirow{2}{*}{Method} & \multicolumn{4}{c}{EndoVis 2018} & \multicolumn{4}{c}{EchoNet-Dynamic} \\ \cline{2-9}
         & 1-click & 3-clicks & 5-clicks & GT-mask & 1-click & 3-clicks & 5-clicks & GT-mask\\
         \hline
         SAM-2 (Hiera-T) & 58.4 & 69.9 & 71.4 & 71.6 & 6.6 & 56.2 & 69.8 & 70.0 \\
         SAM-2 (Hiera-S) & 59.7 & 71.3 & 72.6 & 73.8 & 7.3 & 58.4 & 70.2 & 71.5 \\
         SAM-2 (Hiera-B+) & 60.1 & 70.8 & 73.3 & 74.2 & 10.0 & 65.0 & 71.6 & 72.8 \\
         SAM-2 (Hiera-L) & 60.9 & 71.2 & 74.8 & 77.4 & 7.1 & 65.9 & 67.3 & 73.9 \\
         \hline
    \end{tabular}
    \end{adjustbox}
    \label{tab: jf}
\end{table}

\begin{table}[h]
    \centering
    \small
    \caption{Different-sized models' $\mathcal{J\&F}$ scores ($\%$) under zero-shot conditions over multi-interacted frames under 3-clicks prompt.}
    \begin{adjustbox}{width=0.99\textwidth}
    \begin{tabular}{lccccc|ccccc}
    \hline
        \multirow{2}{*}{Method} & \multicolumn{5}{c}{EndoVis 2018} & \multicolumn{5}{c}{EchoNet-Dynamic} \\ \cline{2-11}
         & 1-frame & 2-frames & 4-frames & 6-frames & 8-frames & 1-frame & 2-frames & 4-frames & 6-frames & 8-frames \\
         \hline
         SAM-2 (Hiera-T) & 69.9 & 69.9 & 72.5 & 74.9 & 76.6 & 56.2 & 57.0 & 63.5 & 68.1 & 71.2 \\
         SAM-2 (Hiera-S) & 71.3 & 71.0 & 73.0 & 75.4 & 77.8 & 58.4 & 58.5 & 64.0 & 69.5 & 70.9  \\
         SAM-2 (Hiera-B+) & 70.8 & 71.2 & 74.1 & 77.6 & 78.2 & 65.0 & 64.9 & 67.9 & 70.0 & 73.4 \\
         SAM-2 (Hiera-L) & 71.2 & 73.0 & 75.2 & 78.0 & 79.9 & 65.9 & 66.1 & 70.3 & 72.0 & 72.5  \\
         \hline
    \end{tabular}
    \end{adjustbox}
    \label{tab: frames}
\end{table}

\subsubsection{Experimental Setup}
In our evaluation of SAM-2 under the condition of zero-shot, we chose Jaccard and F-Score ($\mathcal{J\&F}$) as our evaluation metrics. The Jaccard index describes the Intersection over Union (IoU) between the predicted mask and the ground truth (gt), while the F-Score quantifies the alignment between the boundaries of the predicted mask and the gt boundaries.

\subsubsection{Results}

\begin{figure}[h]
    \centering
    \includegraphics[width=\linewidth]{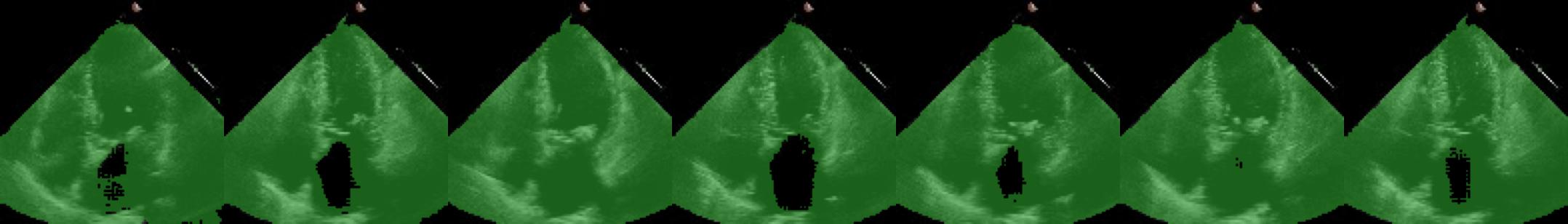}
    \caption{The failure example of the SAM-2 in 1-click applications on low-resolution ultrasound videos.}
    \label{fig: fail ct}
\end{figure}

The most significant breakthrough of SAM-2 compared to SAM is its capability to support video tracking of internal objects. As illustrated in Table~\ref{tab: jf}, we conducted a detailed and comprehensive evaluation of the capabilities of SAM-2 in zero-shot segmentation of medical videos. The results indicate that click counts effectively enhance the accuracy of the outcomes. Just a few additional point prompts enable the model to accurately delineate the object boundaries. Notably, the EchoNet Dynamic~\cite{ouyang2020video} showed the most significant improvements. As shown in Figure~\ref{fig: fail ct}, with only one click, SAM-2 segmented the imaging sector area. However, increasing the number of clicks resulted in a substantial rise in the $\mathcal{J\&F}$ scores from single digits to over 70. Moreover, the model generally performs better with larger sizes when the number of clicks remains the same.

\begin{figure}[h]
    \centering
    \includegraphics[width=\linewidth]{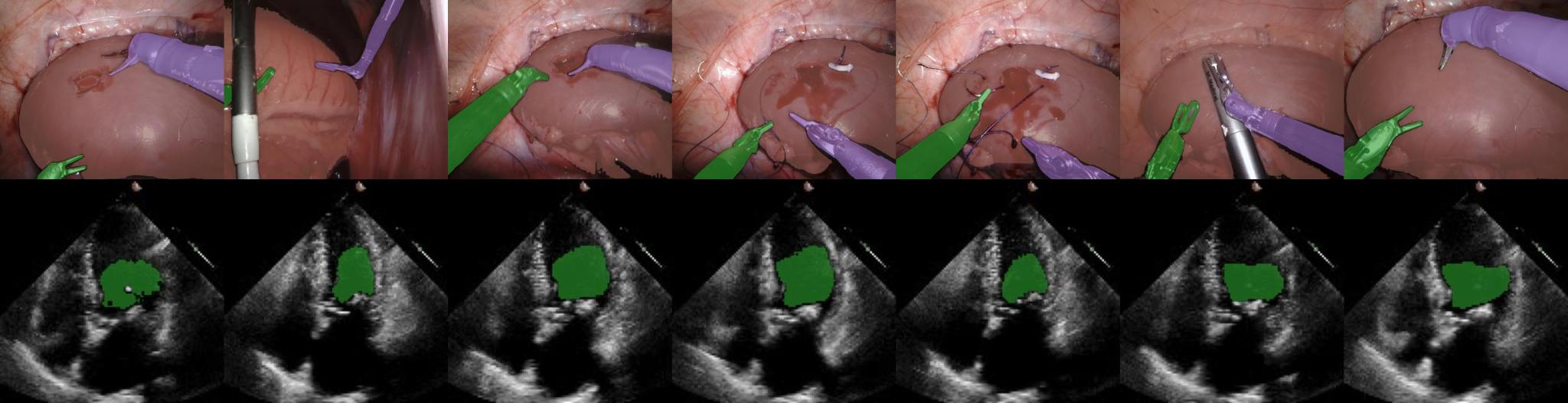}
    \caption{The visualization results of SAM-2 in 3-click applications on two medical scenarios.}
    \label{fig: video}
\end{figure}

A major issue with SAM-2 video segmentation is that if the target object does not appear in the annotated frame, subsequent tracking and segmentation within the video are unfeasible. We addressed this problem by increasing the number of interacted frames. The results in Table~\ref{tab: frames} demonstrate the enhancements achieved by this method. With the addition of interacted frames, SAM-2 showed improved performance under a 3-click scenario, even surpassing results obtained using only the Ground Truth mask on the first frame. We observed that even if the same object was repeatedly marked in the interacted frames, it still enhanced the subsequent segmentation results. We hypothesize that this injection of information allows the model to better recognize the same object from different perspectives, thereby achieving better outcomes. Overall, while SAM-2 demonstrates some capability in zero-shot segmentation of medical videos, its lack of training in medical content renders it somewhat perplexed in certain medical scenarios. This underscores the importance of training BioSAM-2 on video data in subsequent efforts.





\section{Conclusion}
In conclusion, our development of BioSAM-2 represents a significant advancement in biomedical domain. Through the implementation of three specialized evaluation pipelines designed for biomedical images and videos, we have rigorously assessed the performance of SAM-2 across diverse medical scenarios and objects of interest. Our results indicate that the enhanced BioSAM-2 not only outperforms current state-of-the-art foundation methods, but also surpasses the performance of majority specialized models trained specifically for the same medical modalities. These findings affirm the potential of BioSAM-2 as a novel biomedical segmentation approach towards more efficient, accurate, and adaptable diagnostic technologies.

\bibliographystyle{unsrt}

\bibliography{references}

\end{document}